%% file: main.tex
\documentclass[]{aircc}

\usepackage{amsmath}
\usepackage{flexisym}

\usepackage{blindtext}
\usepackage{graphicx}
\usepackage{multirow}
\usepackage{color}
\usepackage[dvipsnames]{xcolor}
\usepackage[normal, justification=centering]{caption}
\usepackage{hyperref}
\usepackage{diagbox}
\usepackage{tikz}
\usepackage{makecell}

\makeatletter
\newcommand{\thickhline}{%
    \noalign {\ifnum 0=`}\fi \hrule height 1pt
    \futurelet \reserved@a \@xhline
}
\newcolumntype{"}{@{\hskip\tabcolsep\vrule width 1pt\hskip\tabcolsep}}
\makeatother

\usepackage{tablefootnote}

\begin{document}

\title{CNN features are also great at unsupervised classification}

\author{Joris~Gu\'erin,
        Olivier~Gibaru,
        St\'ephane Thiery,
        and~Eric~Nyiri}
        
\affiliation{Laboratoire des Sciences de l'Information et des Syst\`emes (CNRS UMR 7296)\\
Arts et M\'etiers ParisTech, Lille, France\\
\small{joris.guerin@ensam.eu}}

\maketitle

\begin{abstract}
  This paper aims at providing insight on the transferability of deep CNN features to unsupervised problems. We study the impact of different pretrained CNN feature extractors on the problem of image set clustering for object classification as well as fine-grained classification. We propose a rather straightforward pipeline combining deep-feature extraction using a CNN pretrained on ImageNet and a classic clustering algorithm to classify sets of images. This approach is compared to state-of-the-art algorithms in image-clustering and provides better results. These results strengthen the belief that supervised training of deep CNN on large datasets, with a large variability of classes, extracts better features than most carefully designed engineering approaches, even for unsupervised tasks. We also validate our approach on a robotic application, consisting in sorting and storing objects smartly based on clustering. 
\end{abstract}


\input{introduction}

\setcounter{subsection}{0}
\setcounter{subsubsection}{0}
\input{methodology}

\setcounter{subsection}{0}
\input{experimental_validation}

\setcounter{subsection}{0}
\setcounter{subsubsection}{0}
\input{application_robustness}

\setcounter{subsection}{0}
\setcounter{subsubsection}{0}
\input{conclusion}

\renewcommand{\thesection}{\hspace*{0em}}
\section{Acknowledgements}
This work is also supported by the European Union's 2020 research and innovation program under grant agreement No.688807, project ColRobot (collaborative robotics for assembly and kitting in smart manufacturing).

The authors would like to thank Jorge Palos, Gil Boye de Sousa, Leon Sixt and Harshvardan Gazula for their precious advices and technical support.

\bibliographystyle{IEEEtran}
\bibliography{biblio}

\section{Authors}
\begin{minipage}{0.8\textwidth}
    \textbf{Joris Gu\'erin} received the dipl\^ome d'ing\'enieur (equivalent to M.Sc. degree) from Arts et M\'etiers ParisTech and the M.Sc. in Industrial Engineering from Texas Tech University, both in 2015. He is currently a Ph.D student at Laboratoire d'Ing\'enierie des Syst\`emes Physiques et Num\'eriques (LISPEN), at Arts et M\'etiers ParisTech, Lille, France. His current research focuses on Transfer Clustering and Computer Vision for Robotics.
\end{minipage}
\begin{minipage}{0.2\textwidth}
    \flushright
    \includegraphics[width=1in,height=1.25in,clip,keepaspectratio]{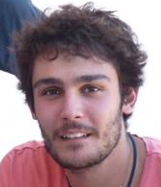}
\end{minipage}

\vspace{15pt}

\begin{minipage}{0.8\textwidth}
    \textbf{Olivier Gibaru} is currently full professor at the Department of Mathematics and Computer Science at ENSAM, Lille campus. He obtained his PhD in applied mathematics in 1997. His main research interests includes: applied mathematics, estimation for robotic applications, geometry, control engineering and high precision mechanical systems. He is the coordinator of the EU Horizon 2020 ColRobot project www.colrobot.eu. He is an active member of the SMAI-SIGMA group which is a national learned society dedicated to Applied Mathematics for the Industrial Applications.
\end{minipage}
\begin{minipage}{0.2\textwidth}
    \flushright
    \includegraphics[width=1in,height=1.25in,clip,keepaspectratio]{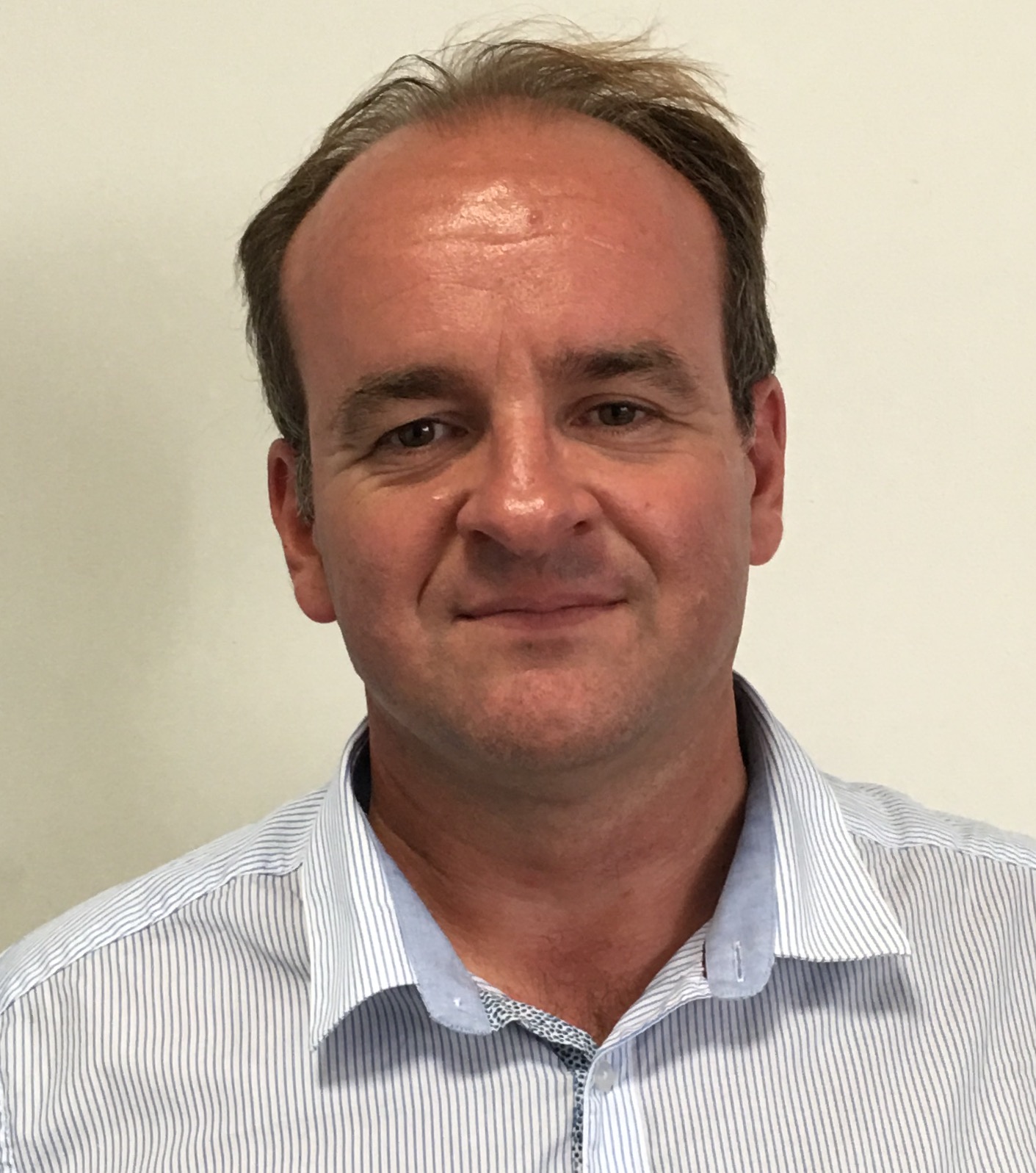}
\end{minipage}

\vspace{15pt}

\begin{minipage}{0.8\textwidth}
    \textbf{St\'ephane Thiery} received the Ph.D degree in Automatics from University of Nice-Sophia Antipolis, France, in 2008. He was a post-doctoral fellow in the NON-A team in INRIA-Lille, for eight months in 2009-2010, and joined Arts et M\'etiers ParisTech Engineering School, as assistant professor in Applied Mathematics and Automatics, in 2010. His current research includes machine learning, real-time parameters estimation, and control of mechanical systems.
\end{minipage}
\begin{minipage}{0.2\textwidth}
    \flushright
    \includegraphics[width=1in,height=1.25in,clip]{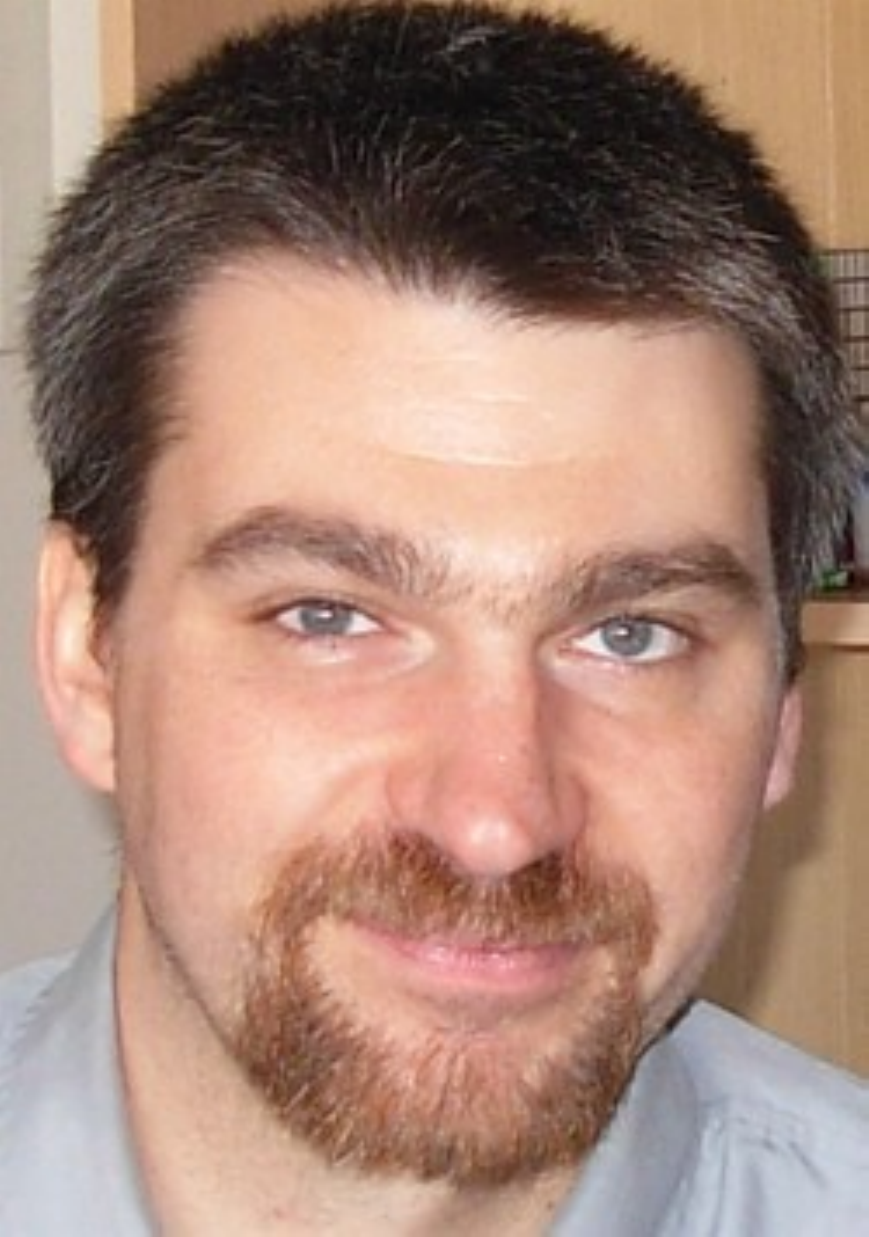}
\end{minipage}

\vspace{15pt}

\begin{minipage}{0.8\textwidth}
    \textbf{Eric Nyiri} received the Ph.D degree in Computer Science from University of Lille I, France, in 1994. He joined Arts et M\'etiers ParisTech Engineering School, as an assistant professor in Applied Mathematics and Computer Science, in 1995. His initial research domain was L1 interpolation and approximation. In 2010, he joined the LSIS Lab and his current research includes machine learning and path planning for robots. Since 2016, he is a member of the COLROBOT European project.
\end{minipage}
\begin{minipage}{0.2\textwidth}
    \flushright
    \includegraphics[width=1in,height=1.25in,clip]{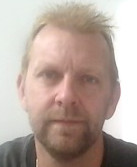}
\end{minipage}

\end{document}

%% file: introduction.tex
\section{Introduction}

In a close future, it is likely to see industrial robots performing tasks requiring to make complex decisions. In this perspective, we have developed an automatic sorting and storing application (see section 1.1.2) consisting in clustering images based on semantic content and storing the objects in boxes accordingly using a serial robot (\url{https://youtu.be/NpZIwY3H-gE}). This application can have various uses in shopfloors (workspaces can be organized before the workday, unsorted goods can be sorted before packaging, ...), which motivated this study of image-set clustering.

As of today, deep convolutional neural networks (CNN) \cite{DL_lecun} are the method of choice for supervised image classification. Since \cite{alexnet} demonstrated astounding results on ImageNet, all other methods have rapidly been abandoned for ILSVRC \cite{imagenet}. As suggested by \cite{largeDS}, performances of CNN are highly correlated to the amount of labeled training data available. Nevertheless, even when few labels are available, several recent papers \cite{feat_extrac1, feat_extrac2, feat_extrac3} have shown that CNN can still outperform any other approach by transferring knowledge learned on large datasets. In particular, \cite{feat_extrac1} has shown that extracting features from OverFeat \cite{overfeat} pretrained on ImageNet, and training a simple Support Vector Machine (SVM) \cite{pattern_rec} classifier on these features to fit the new dataset provides better results than many more complex approaches for supervised classification. These results demonstrate that a CNN, trained on a large and versatile dataset, learns information about object characteristics that is generic enough to transfer to objects that are not in the original dataset. 

While developing the automatic robotic sorting and storing application, we needed to classify sets of images based on their content, in an unsupervised way. Multiple papers introduced methods to solve unsupervised object classification from sets of images (see section 1.1.1), producing relatively good results. However, we wanted to know if the information from a large and versatile dataset, stored in the weights of a CNN, could be used straightforwardly to outperform state-of-the-art algorithms at unsupervised image-set classification. The goal of this paper is to answer the following question: \emph{How good are features extracted with a CNN pretrained on a large dataset, for unsupervised image classification tasks?} To do so, we use a similar approach to \cite{feat_extrac1}, consisting in applying classic clustering methods to features extracted from the final layers of a CNN (see section 2 for more details) and comparing it with state-of-the-art image set clustering algorithms \cite{infinite_ensemble, commonality_clustering} on several public datasets. 

The intuition behind such approach for unsupervised object classification is that, as it works with SVM \cite{feat_extrac1}, the CNN must project data in a feature space where they are somehow linearly separable. Thus, simple clustering algorithms such as K-means might be working well. However, this study is interesting as the performance of such simple clustering algorithms often depends on the notion of distance between points, on which we remain uncertain.

\subsection{Previous work}
\subsubsection
  {Image-set clustering}
  \label{sec:prev_im_clust}
Given a set of unlabeled images, the image-set clustering problem consists in finding subsets of images based on their content: two images representing the same object should be clustered together and separated from images representing other objects. Figure \ref{fig:appli} illustrates the expected output from an image-set clustering algorithm in the case of our robotics application. This problem should not be confused with image segmentation \cite{segmentation}, which is also sometimes called image clustering. 

Image-set clustering has been widely studied for two decades. It has applications for searching large image database \cite{webscale1, webscale2, webscale3}, concept discovery in images \cite{conceptdiscovery}, storyline reconstruction \cite{storyline}, medical images classification \cite{alternating_opt_clust}, ... The first successful methods focused on feature selection and used sophisticated algorithms to deal with complex features. For instance, \cite{GMM} represents images by Gaussian Mixture Models fitted to the pixels and clusters the set using the Information Bottleneck (IB) method \cite{info_bottleneck}. \cite{segmentation_image_set} uses features resulting from image joint segmentation and sequential IB for clustering. \cite{commonality_clustering} uses Bags of Features with local representations (SIFT, SURF, ...) and defines commonality measures used for agglomerative clustering. Recently, image-set clustering algorithms have shifted towards using deep features. \cite{infinite_ensemble}  uses deep auto-encoders combined with ensemble clustering to generate feature representations suitable for clustering. \cite{jule, alternating_opt_clust} learns jointly the clusters and the representation using alternating optimization \cite{alter_opt_bezdek}.

\subsubsection
  {Robotic automatic sorting application}
  \label{sec:robot_appli}
  
\begin{figure}[t]
    \centering
    
    \begin{tikzpicture}
    \node[inner sep=0pt] (im0) at (0,0)
        {\includegraphics[width=.1\textwidth]{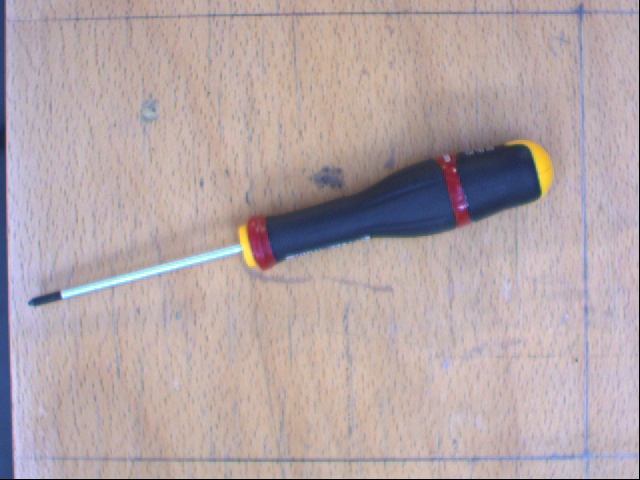}};
    \node[inner sep=0pt] (im1) at (0, -1.20)
        {\includegraphics[width=.1\textwidth]{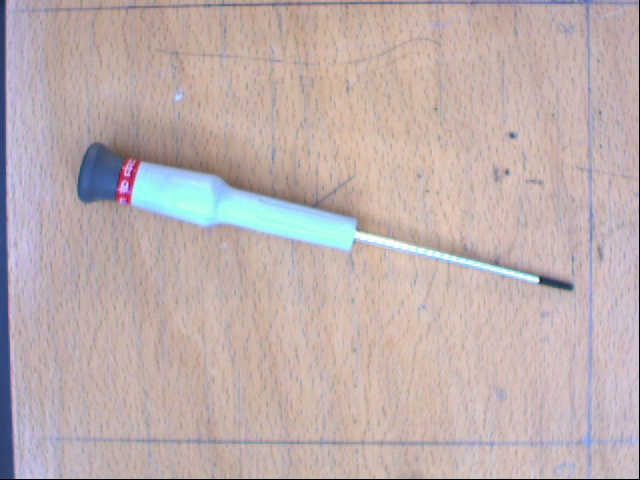}};
    \node[inner sep=0pt] (im2) at (0, -2.40)
        {\includegraphics[width=.1\textwidth]{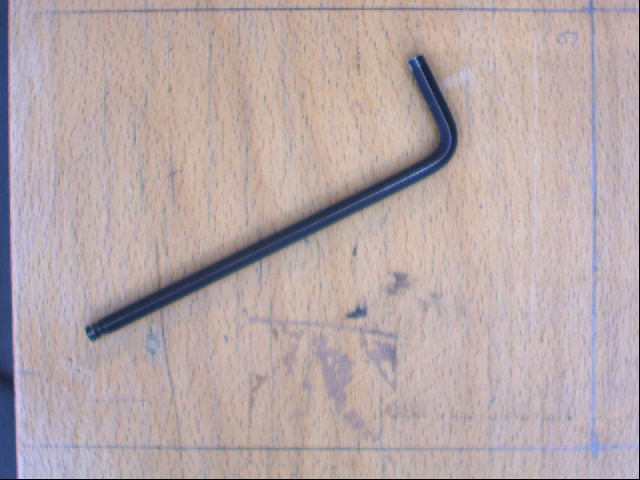}};
    \node[inner sep=0pt] (im3) at (1.57, 0)
        {\includegraphics[width=.1\textwidth]{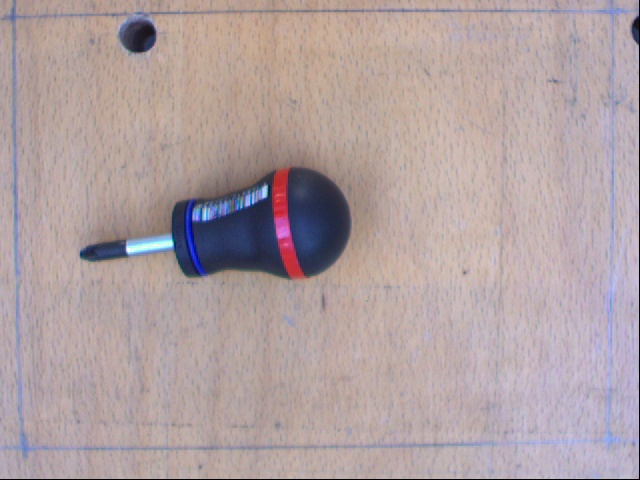}};
    \node[inner sep=0pt] (im4) at (1.57, -1.20)
        {\includegraphics[width=.1\textwidth]{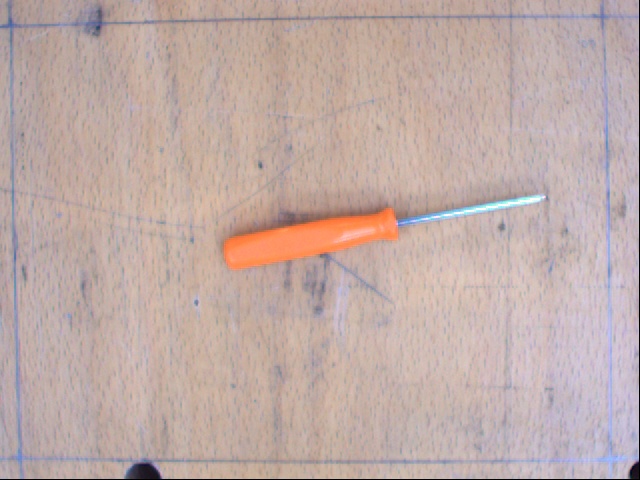}};
    \node[inner sep=0pt] (im5) at (1.57, -2.40)
        {\includegraphics[width=.1\textwidth]{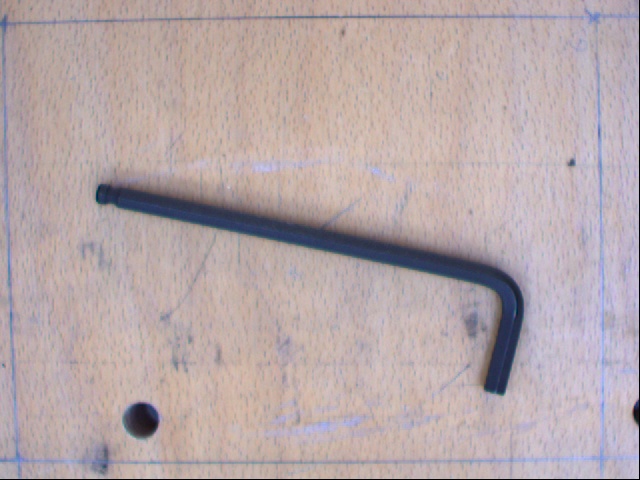}};
    \node[inner sep=0pt] (im6) at (3.14, 0)
        {\includegraphics[width=.1\textwidth]{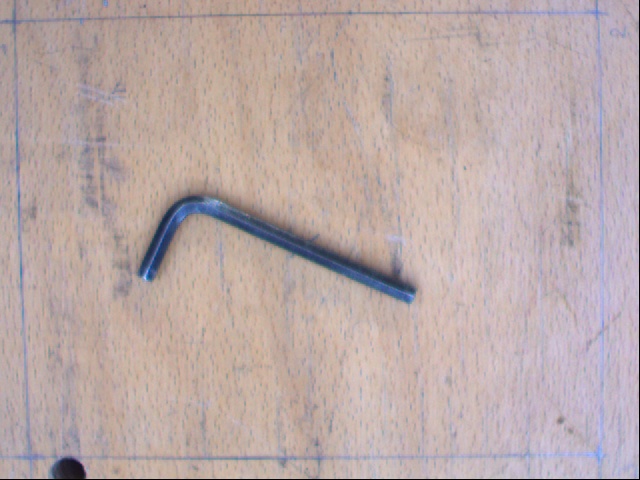}};
    \node[inner sep=0pt] (im7) at (3.14, -1.20)
        {\includegraphics[width=.1\textwidth]{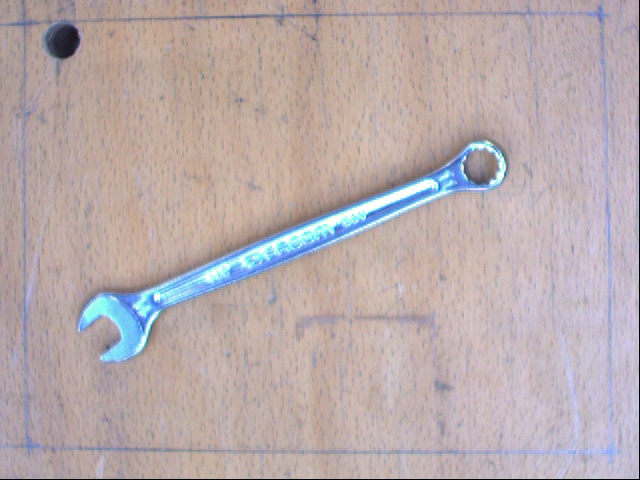}};
    \node[inner sep=0pt] (im8) at (3.14, -2.40)
        {\includegraphics[width=.1\textwidth]{image_3.jpg}};
    \node[inner sep=0pt] (out) at (10, -1.20)
        {\includegraphics[width=.4\textwidth]{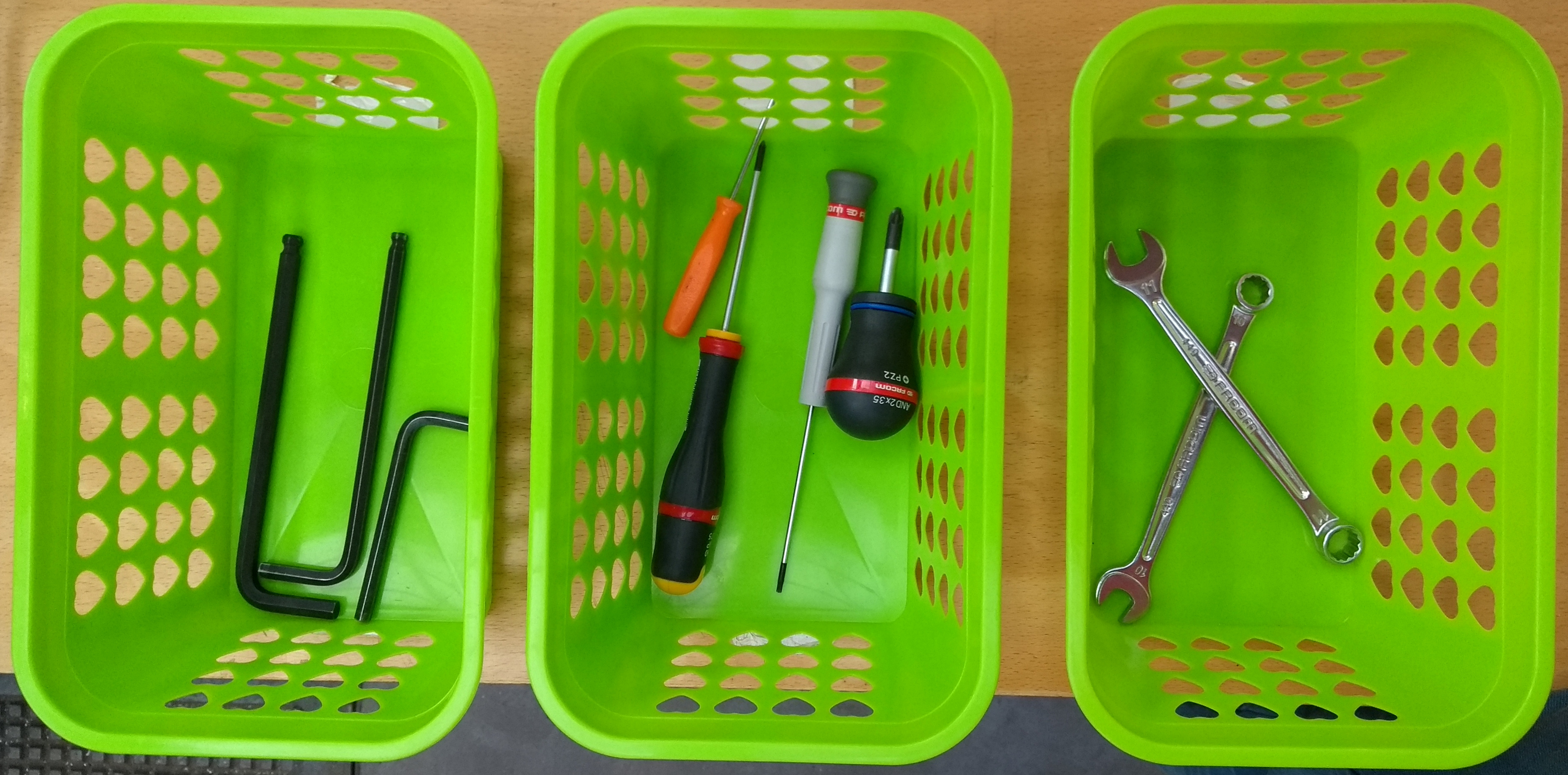}};
    \draw[->,line width=0.5mm] (im7.east) -- (out.west)
        node[above, midway] {Clustering} node[below, midway] {+ Robot sorting};
    \end{tikzpicture}
    
    \caption{Robotic application description}
    \label{fig:appli}
\end{figure}

\setcounter{footnote}{0}

Together with showing that \textit{deep features + simple clustering} outperforms other approaches on unsupervised object classification, we apply this pipeline to solve an automatic smart robot sorting application first introduced in \cite{gr_kmeans}. The idea is that the robot, equipped with a camera, visualizes a set of objects placed on a table. Given the number of storage boxes available, it needs to figure out the best way of sorting and storing the objects before physically doing it. The approach proposed in this paper exploits semantic information (deep features) while \cite{gr_kmeans} uses a computer vision algorithm to extract shapes and colors. A video of the application can be seen at (\url{https://youtu.be/NpZIwY3H-gE}). An example of inputs/output of the application is shown in Figure \ref{fig:appli}. The robustness of this application is also investigated in section 4 by changing the lighting conditions, the position and orientation of the objects as well as the background. For this robustness validation, we built a dataset that appears to be a challenging one for image-set clustering (\url{https://github.com/jorisguerin/toolClustering_dataset}).

\subsection{Contributions}

The main contribution of this paper is to convey further insight into deep CNN features by showing their scalability to unsupervised classification problems. We also propose a new baseline on several image clustering tasks. 

Other contributions include the implementation of an application combining unsupervised image classification with robotic sorting. The method proposed to solve this problem, constitutes a new step towards autonomous decision-making robots. The dataset introduced in this paper, which is relatively challenging for image-set clustering, is also a contribution that can be used in further research to investigate robustness to background and lighting conditions for image clustering algorithms.

%% file: methodology.tex
\section
  {Clustering images with deep feature extraction}
  \label{sec:pipeline}

\subsection
  {Pipeline description}
  
The pipeline we propose for image set clustering is fairly straightforward. It consists in extracting deep features from all the images in the set, by using a deep convolutional neural network pretrained on a large dataset for image classification and then apply a "standard" clustering algorithm to these features. We initially tried this approach as a first step towards developing a better clustering algorithm, however, it appears that this simple approach outperforms state-of-the-art algorithm at image-set clustering.

To implement this unsupervised image classification pipeline, we first need to answer four questions:
\begin{itemize}
    \item What dataset should be used for pretraining?
    \item What CNN architecture should we use?
    \item Which layer output should be chosen for feature extraction?
    \item What clustering algorithm should be used?
\end{itemize}

As of today, ImageNet is the only very large labelled public dataset which has enough variety in its classes to be a good feature extractor for a variety of tasks. Moreover, there are plenty of CNN pretrained on ImageNet already available online. Hence, we will use a CNN pretrained on ImageNet. The three other questions are answered experimentally. We use the VOC2007 \cite{voc2007} test set without labels, which is new for the pretrained net, to compare performances of the different options.

To ease development and accelerate implementation, we compare the Keras \cite{keras} implementations of ResNet50 \cite{resnet}, InceptionV3 \cite{inception}, VGG16, VGG19 \cite{vgg} and Xception \cite{xception} with the pretrained weights provided by Keras. For the clustering algorithms, we use the scikit-learn \cite{sklearn} implementations of K-means (KM) \cite{kmeans++}, Minibatch K-means (MBKM) \cite{mbkm}, Affinity Propagation (AP) \cite{affinity}, Mean Shift (MS) \cite{mean_shift}, Agglomerative Hierarchical Clustering (AC) \cite{agglomerative}, DBScan (DBS) \cite{dbscan} and Birch (Bi) \cite{birch}. For each CNN, the layers after which the features are extracted can be found in Table \ref{table_result} (Layers names are the same as in the Keras implementations).

In the image-set clustering problem, the expected classes are represented by objects present on the picture and for this reason we need semantic information, which is present in the final layers of the network. Thus, we only choose layers among the last layers of the networks. On the one hand, the last one or two layers might provide better results as their goal is to separate the data (at least for the fully-connected layers). On the other hand, the opposite intuition is also relevant as we can imagine that these layers are too specialized to be transferable. These two contradictory arguments motivated the following experiment.

We also note that the test set of VOC2007 has been modified for this validation. We removed all the images presenting two or more labels in order to have ground truth to compute Normalized Mutual Information (NMI) scores. Indeed, if an image possesses several labels we cannot judge if the clustering pipeline classified it properly or not. We note VOC2007-SL (single label) the modified VOC2007 test set.

\subsection
  {Hyperparameters choice}
  \label{sec:hyperparam}
 
To answer the questions stated above, we try to cluster the VOC2007-SL set using all combinations of CNN architectures, layer choices and clustering algorithms. To compare performances, we use NMI scores. We also report clustering time for completeness. Only scikit-learn default hyperparameters of the different clustering algorithms are used, which illustrate the simplicity of this approach. For KM and MBKM, as the results depend on random initialization, experiments are run ten times and reported results are averaged over the different runs.

\input{table_res_comp.tex}

Looking at the results, we choose Agglomerative Clustering on features extracted from the final layer of an Xception CNN pretrained on ImageNet for image-set clustering. This pipeline is then compared to state-of-the-art methods in the next section in order to see how transferable CNN ImageNet features are for unsupervised categorization.

%% file: table_res_comp.tex
\begin{table}[!ht]
\caption{NMI scores (in black) and time in seconds (in blue, italics) on Pascal VOC2007-SL test set using different CNN, different output layers and different clustering algorithms. (Layers names are the same as in the Keras implementations).}
\label{table_result}
\centering
\vspace{\baselineskip}
    \begin{tabular}{c|c"c|c|c|c|c|c|c}
        
        \multicolumn{2}{c"}{} & KM & MBKM & AP & MS & AC & DBS \tablefootnote{The poor results with DB-Scan might come from the default parameters. We might get better results using different configurations, but this is out of the scope of this paper.} & Bi\tabularnewline
        \Xhline{2\arrayrulewidth}
        \multirow{6}{*}{Inception V3} & \multirow{2}{*}{mixed9} & 0.108 & 0.105 & 0.219 & 0.153 & 0.110 & 0 & 0.110 \tabularnewline 
        & & \textit{\textcolor{Blue}{374}} & \textit{\textcolor{Blue}{7.3}} & \textit{\textcolor{Blue}{12.7}} & \textit{\textcolor{Blue}{16281}} & \textit{\textcolor{Blue}{525}} & \textit{\textcolor{Blue}{138}} & \textit{\textcolor{Blue}{577}} \tabularnewline \cline{2-9}
        
        & \multirow{2}{*}{mixed10} & 0.468 & 0.401 & 0.442 & 0.039 & 0.595 & 0 & 0.595 \tabularnewline 
        & & \textit{\textcolor{Blue}{609}} & \textit{\textcolor{Blue}{5.1}} & \textit{\textcolor{Blue}{8.5}} & \textit{\textcolor{Blue}{12126}} & \textit{\textcolor{Blue}{525}} & \textit{\textcolor{Blue}{119}} & \textit{\textcolor{Blue}{567}} \tabularnewline \cline{2-9}
        
        & \multirow{2}{*}{avg_pool} & 0.674 & 0.661 & 0.621 & 0.024 & 0.686 & 0 & 0.686 \tabularnewline 
        & & \textit{\textcolor{Blue}{6.3}} & \textit{\textcolor{Blue}{0.2}} & \textit{\textcolor{Blue}{7.7}} & \textit{\textcolor{Blue}{230}} & \textit{\textcolor{Blue}{8.5}} & \textit{\textcolor{Blue}{1.8}} & \textit{\textcolor{Blue}{8.9}} \tabularnewline \hline
        
        \multirow{2}{*}{Resnet 50} & \multirow{2}{*}{avg_pool} & 0.6748 & 0.641 & 0.587 & 0.043 & 0.640 & 0 & 0.640 \tabularnewline 
        & & \textit{\textcolor{Blue}{7.0}} & \textit{\textcolor{Blue}{0.1}} & \textit{\textcolor{Blue}{4.6}} & \textit{\textcolor{Blue}{197}} & \textit{\textcolor{Blue}{8.0}} & \textit{\textcolor{Blue}{1.9}} & \textit{\textcolor{Blue}{8.9}} \tabularnewline \hline
        
        \multirow{8}{*}{VGG 16} & \multirow{2}{*}{block4_pool} & 0.218 & 0.085 & 0.133 & 0.124 & 0.277 & 0 & 0.277 \tabularnewline 
        & & \textit{\textcolor{Blue}{278}} & \textit{\textcolor{Blue}{3.6}} & \textit{\textcolor{Blue}{6.0}} & \textit{\textcolor{Blue}{10010}} & \textit{\textcolor{Blue}{391}} & \textit{\textcolor{Blue}{82.8}} & \textit{\textcolor{Blue}{436}} \tabularnewline \cline{2-9}
        
        & \multirow{2}{*}{block5_pool} & 0.488 & 0.048 & 0.262 & 0.194 & 0.530 & 0 & 0.530 \tabularnewline 
        & & \textit{\textcolor{Blue}{78}} & \textit{\textcolor{Blue}{1.1}} & \textit{\textcolor{Blue}{9.3}} & \textit{\textcolor{Blue}{2325}} & \textit{\textcolor{Blue}{99}} & \textit{\textcolor{Blue}{21}} & \textit{\textcolor{Blue}{107}} \tabularnewline \cline{2-9}
        
        & \multirow{2}{*}{fc1} & 0.606 & 0.458 & 0.421 & 0.187 & 0.617 & 0 & 0.617 \tabularnewline 
        & & \textit{\textcolor{Blue}{17}} & \textit{\textcolor{Blue}{0.2}} & \textit{\textcolor{Blue}{4.8}} & \textit{\textcolor{Blue}{365}} & \textit{\textcolor{Blue}{17}} & \textit{\textcolor{Blue}{3.8}} & \textit{\textcolor{Blue}{19}} \tabularnewline \cline{2-9}
        
        & \multirow{2}{*}{fc2} & 0.661 & 0.611 & 0.551 & 0.085 & 0.673 & 0 & 0.673 \tabularnewline 
        & & \textit{\textcolor{Blue}{16}} & \textit{\textcolor{Blue}{0.2}} & \textit{\textcolor{Blue}{4.3}} & \textit{\textcolor{Blue}{373}} & \textit{\textcolor{Blue}{15.9}} & \textit{\textcolor{Blue}{3.8}} & \textit{\textcolor{Blue}{19.7}} \tabularnewline \hline
        
        \multirow{8}{*}{VGG 19} & \multirow{2}{*}{block4_pool} & 0.203 & 0.139 & 0.124 & 0.135 & 0.234 & 0 & 0.234 \tabularnewline 
        & & \textit{\textcolor{Blue}{220}} & \textit{\textcolor{Blue}{3.7}} & \textit{\textcolor{Blue}{6.3}} & \textit{\textcolor{Blue}{10298}} & \textit{\textcolor{Blue}{388}} & \textit{\textcolor{Blue}{83}} & \textit{\textcolor{Blue}{435}} \tabularnewline \cline{2-9}
        
        & \multirow{2}{*}{block5_pool} & 0.522 & 0.321 & 0.250 & 0.198 & 0.540 & 0 & 0.540 \tabularnewline 
        & & \textit{\textcolor{Blue}{74}} & \textit{\textcolor{Blue}{0.9}} & \textit{\textcolor{Blue}{9.3}} & \textit{\textcolor{Blue}{2353}} & \textit{\textcolor{Blue}{97}} & \textit{\textcolor{Blue}{20}} & \textit{\textcolor{Blue}{106}} \tabularnewline \cline{2-9}
        
        & \multirow{2}{*}{fc1} & 0.607 & 0.471 & 0.449 & 0.188 & 0.628 & 0 & 0.628 \tabularnewline 
        & & \textit{\textcolor{Blue}{17}} & \textit{\textcolor{Blue}{0.2}} & \textit{\textcolor{Blue}{9.3}} & \textit{\textcolor{Blue}{365}} & \textit{\textcolor{Blue}{17}} & \textit{\textcolor{Blue}{3.9}} & \textit{\textcolor{Blue}{18}} \tabularnewline \cline{2-9}
        
        & \multirow{2}{*}{fc2} & 0.672 & 0.615 & 0.557 & 0.083 & 0.671 & 0 & 0.671 \tabularnewline 
        & & \textit{\textcolor{Blue}{15}} & \textit{\textcolor{Blue}{0.2}} & \textit{\textcolor{Blue}{5.6}} & \textit{\textcolor{Blue}{391}} & \textit{\textcolor{Blue}{17}} & \textit{\textcolor{Blue}{3.9}} & \textit{\textcolor{Blue}{18}} \tabularnewline \hline
        
        \multirow{6}{*}{Xception} & \multirow{2}{*}{block13_pool} & 0.376 & 0.264 & 0.351 & 0.044 & 0.473 & 0 & 0.473 \tabularnewline 
        & & \textit{\textcolor{Blue}{410}} & \textit{\textcolor{Blue}{4.8}} & \textit{\textcolor{Blue}{10}} & \textit{\textcolor{Blue}{9677}} & \textit{\textcolor{Blue}{403}} & \textit{\textcolor{Blue}{87}} & \textit{\textcolor{Blue}{444}} \tabularnewline \cline{2-9}
        
        & \multirow{2}{*}{block14_act} & 0.574 & 0.428 & 0.584 & 0.071 & 0.634 & 0 & 0.634 \tabularnewline 
        & & \textit{\textcolor{Blue}{935}} & \textit{\textcolor{Blue}{10}} & \textit{\textcolor{Blue}{10}} & \textit{\textcolor{Blue}{24809}} & \textit{\textcolor{Blue}{820}} & \textit{\textcolor{Blue}{180}} & \textit{\textcolor{Blue}{901}} \tabularnewline \cline{2-9}
        
        & \multirow{2}{*}{avg_pool} & 0.692 & 0.636 & 0.636 & 0.052 & \textbf{0.726} & 0 & 0.726 \tabularnewline 
        & & \textit{\textcolor{Blue}{7.1}} & \textit{\textcolor{Blue}{0.1}} & \textit{\textcolor{Blue}{4.9}} & \textit{\textcolor{Blue}{201}} & \textit{\textcolor{Blue}{\textbf{8.5}}} & \textit{\textcolor{Blue}{5.5}} & \textit{\textcolor{Blue}{9.1}} \tabularnewline \hline
        
    \end{tabular}
    
   
    

   
\end{table}

%% file: experimental_validation.tex
\section{Validation on several public datasets}

\subsection{Datasets description}

The efficiency of the proposed method is demonstrated by comparing it to other recent successful methods (see section 3.2) on several public datasets which characteristics are described in Table \ref{table_data}. 

\input{table_datasets.tex}

The clustering tasks involved by these datasets are different from each others (Face recognition, grouping different objects, recognizing different pictures of the same object). In addition, the content of the classes differs from the ones in ImageNet. For these reasons, the four datasets constitute a good benchmark to quantify the robustness of transfer learning for unsupervised object categorization.

\subsection{Results comparison}
\label{sec:res_comp}

We propose a comparison with the results reported in the following papers dealing with image set clustering:
\begin{itemize}
    \item \cite{commonality_clustering} proposes different clustering algorithms applied on bags of features. In Table \ref{table_compare}, we note "BoF" the best results obtained by such pipeline on the different datasets.
    
    \item \cite{infinite_ensemble} proposes a method called infinite ensemble clustering (IEC). In the paper, IEC algorithm is compared to several other deep clustering algorithms and ensemble clustering algorithms. In Table \ref{table_compare}, we report the best results obtained using Deep Clustering (DC) and Ensemble Clustering (EC) for each datasets. We note that for VOC2007-5-ML, \cite{infinite_ensemble} also uses deep features as clustering inputs (the CNN used is not reported).
    
    \item \cite{jule} proposes a method called Joint Unsupervised Learning (JULE) of Deep Representations and Image Clusters, based on Alternating optimization between clustering and weight optimization of the CNN feature extractor. Results from this work are reported in Table \ref{table_compare}.
\end{itemize}

For each dataset groundtruth is known as they are intended for supervised classification. We compute both NMI scores and purity for each dataset/method pair.

\input{table_litterature_compare.tex}

Table \ref{table_compare} shows that features extracted by the final layer of Xception combined with Agglomerative Clustering performs better than or close to state-of-the-art methods at unsupervised object classification as well as fine-grained image clustering (ORL). Results on the ORL dataset are interesting as they show that pretrained Xception is able to classify different faces without supervision, although ImageNet does not deal with human faces at all.

This is an important result as it shows that, \emph{with today's methods, given an unlabeled image-set, we can extract more information from a large labeled dataset, with a large variety of classes, than from the set itself}. It is better than hand-engineered approaches (BoF) as well as unsupervised trained deep networks (Deep clustering and Ensemble clustering). It also raises the question of how the representation learning problem should be handled. Indeed, although less satisfactory from a research perspective, it might be more appropriate to work on the creation of a larger database and train networks on it so that it can be used as a knowledge base for many other problems.

We underline the very good results of JULE (\cite{jule}) at clustering COIL100. Regarding the results of this methodology on Scene clustering \cite{alternating_opt_clust}, it appears that fine tuning feature extraction using alternating optimization is a good way of improving clustering results. However, the simple approach proposed here still keeps the advantage of being very fast (as it only requires to evaluate the network once for each sample and apply AC), which is useful for our application for instance.

It is also interesting to notice that \cite{infinite_ensemble} is also using CNN features for clustering VOC2007-5-ML. The fact that our pipeline outperform largely DC and EC for this dataset illustrate that when using transfer learning for a new problem, careful selection of the CNN used is important, which is often neglected.

\subsection{A word on scene clustering}
\label{sec:scene}

The problem studied in this paper is the one raised by the robotic application, unsupervised objects sorting. For this reason, we compared our result at different object categorization tasks as well as fine-grained classification for clustering of similar objects. Another interesting image classification problem is the one of scene clustering, studied in \cite{alternating_opt_clust} on two datasets (\cite{mit67, archi25}). For this task, the pipeline proposed in this paper cannot perform as well as for object classification. Indeed, ImageNet does not contain any class which requires to group objects together. Thus, although the features are good at supervised scene classification, without further training they are not able to group objects together as such behaviour is not encoded in the final layers of the CNN. However, this issue is not inherent to the method defined and we believe that with bigger and more versatile datasets, results would be as good as any other method. 

%% file: table_datasets.tex
\begin{table}[!ht]
\caption{Several key features about the datasets used for method validation.} 
\label{table_data}
\centering

    \begin{tabular}{c|c|c|c|c}
        & Problem type & Image Size & \# Classes & \# Instances\tabularnewline \hline
        COIL100 \cite{coil100} & Object recognition & $128 \times 128$ & 100 & 7201 \tabularnewline \hline
        Nisters \cite{nister} & Object recognition & $640 \times 480$ & 2550 & 10200 \tabularnewline \hline
        ORL \cite{orl} & Face recognition & $92 \times 112$ & 40 & 400 \tabularnewline \hline
        VOC2007-5-ML \tablefootnote{The data used for VOC2007 in \cite{infinite_ensemble} are irrelevant for clustering with hard assignment. The VOC2007 subset used in \cite{infinite_ensemble} contains images with several objects but only one label. However, we still ran our clustering method on this data to be able to compare results. This second modified VOC2007 set is denoted VOC2007-5-ML (5 classes - multiple labels)} & Object recognition & variable & 5 & 3376
    \end{tabular}

\vspace*{3pt}
   
\end{table}

%% file: table_litterature_compare.tex
\begin{table}[!ht]
\caption{NMI scores and purity comparison on various public datasets. (\footnotesize{A result that is not reported in the papers cited above is denoted N.R.})} 
\label{table_compare}
\centering

    \begin{tabular}{c|c|c|c|c|c}
        \multicolumn{6}{c}{NMI scores}\vspace{1mm}\tabularnewline
        & DC & EC & BoF & JULE & Ours (Xception + AC)\tabularnewline \hline
        COIL100 & 0.779 & 0.787 & N.R. & 0.985 & 0.951 \tabularnewline \hline
        Nisters & N.R. & N.R. & 0.918 & N.R. & 0.991 \tabularnewline \hline
        ORL & 0.777 & 0.805 & 0.878 & N.R. & 0.93 \tabularnewline \hline
        VOC2007-5-ML & 0.265 & 0.272 & N.R. & N.R. & 0.367
    \end{tabular}
    
    \vspace*{10pt}
    
    \begin{tabular}{c|c|c|c}
        \multicolumn{4}{c}{Purity}\vspace{1mm}\tabularnewline
        & DC & EC & Ours (Xception + AC)\tabularnewline \hline
        COIL100 & 0.535 & 0.546 & 0.882 \tabularnewline \hline
        Nisters & N.R. & N.R. & 0.988 \tabularnewline \hline
        ORL & 0.578 & 0.630 & 0.875 \tabularnewline \hline
        VOC2007-5-ML & 0.513 & 0.536 & 0.622
    \end{tabular}

\vspace*{3pt}
   
\end{table}

%% file: application_robustness.tex
\section{Application validation}
    \label{sec:robustness}

In the setting tested initially (Figure \ref{fig:appli}), where the set of objects to cluster is composed of screw drivers, flat keys, allen keys and clamps, and where the background is a table, the success rate of the application described in section 1.1.2 is $100\%$. Although for certain classes (flat and allen keys) the intra-cluster similarity is high, it is not the case for the two others. This task is also difficult to solve because we carried out the experiments in a shopfloor under unmastered lighting. 

\begin{figure}[t]
    \centering
    
    \begin{tikzpicture}
    \node[inner sep=0pt] (im0) at (0, 0)
        {\includegraphics[width=.15\textwidth]{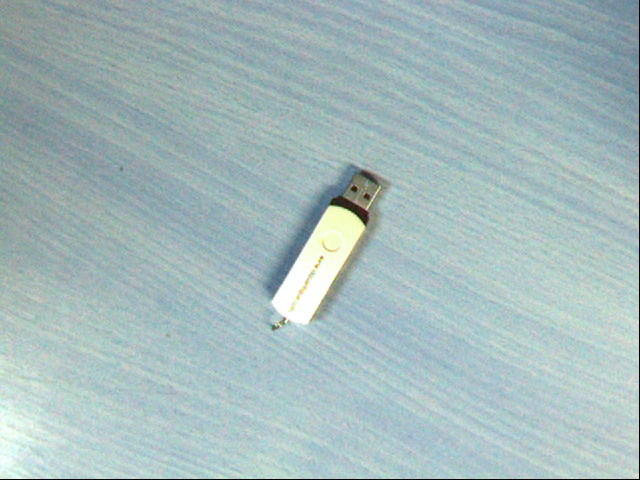}};
    \node[inner sep=0pt] (im1) at (2.35, 0)
        {\includegraphics[width=.15\textwidth]{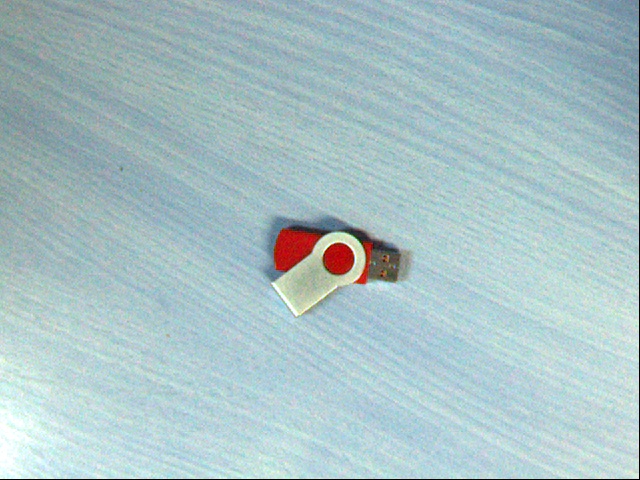}};
    \node[inner sep=0pt] (im2) at (4.7, 0)
        {\includegraphics[width=.15\textwidth]{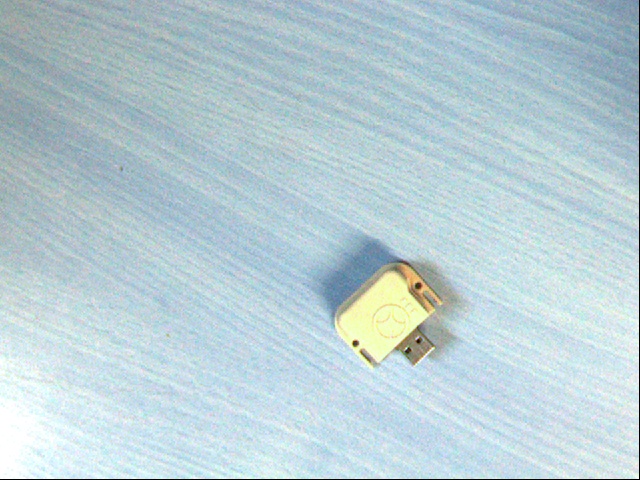}};
    \node[inner sep=0pt] (im3) at (7.2, 0)
        {\includegraphics[width=.15\textwidth]{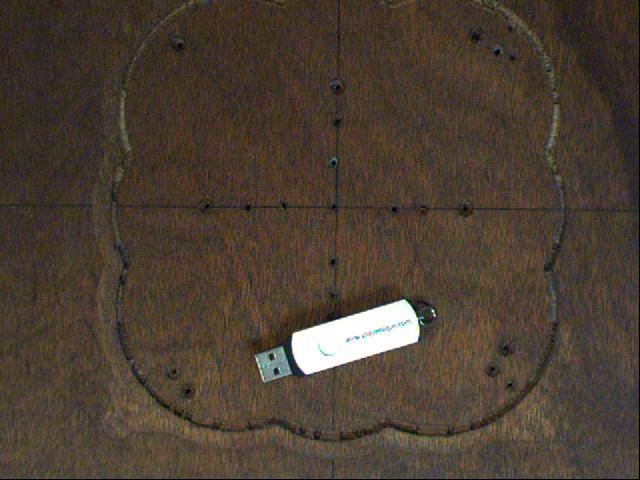}};
    \node[inner sep=0pt] (im4) at (9.55, 0)
        {\includegraphics[width=.15\textwidth]{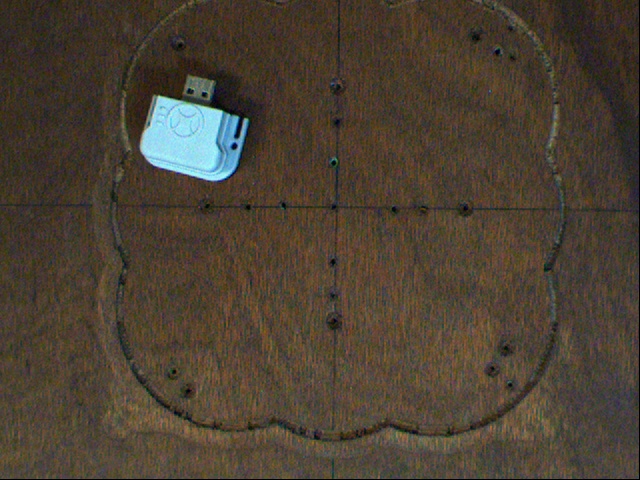}};
    \node[inner sep=0pt] (im5) at (11.9, 0)
        {\includegraphics[width=.15\textwidth]{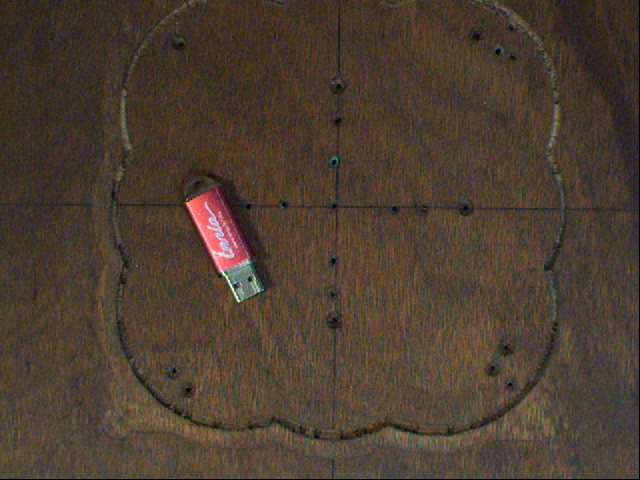}};
    \node[inner sep=0pt] (im6) at (0, -1.8)
        {\includegraphics[width=.15\textwidth]{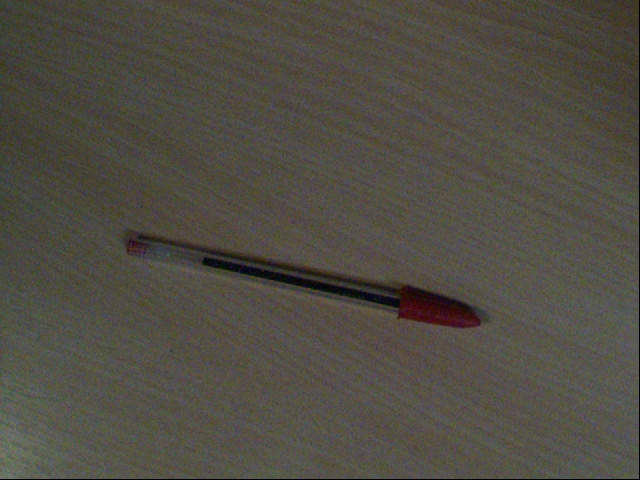}};
    \node[inner sep=0pt] (im7) at (2.35, -1.8)
        {\includegraphics[width=.15\textwidth]{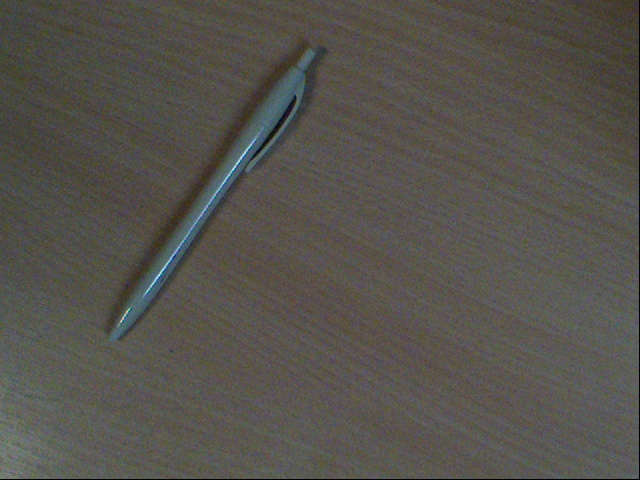}};
    \node[inner sep=0pt] (im8) at (4.7, -1.8)
        {\includegraphics[width=.15\textwidth]{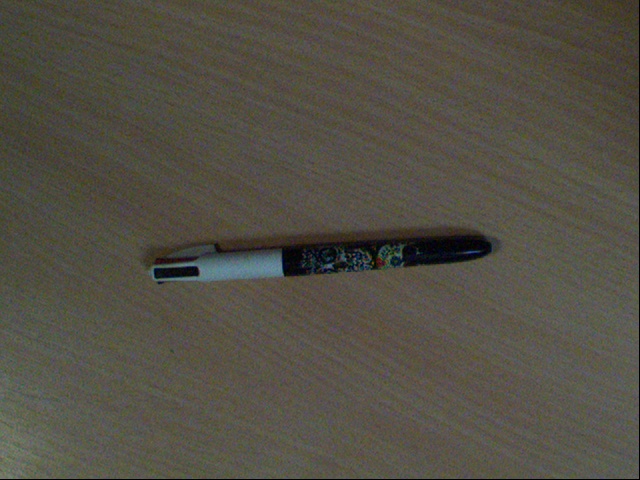}};
    \node[inner sep=0pt] (im9) at (7.2, -1.8)
        {\includegraphics[width=.15\textwidth]{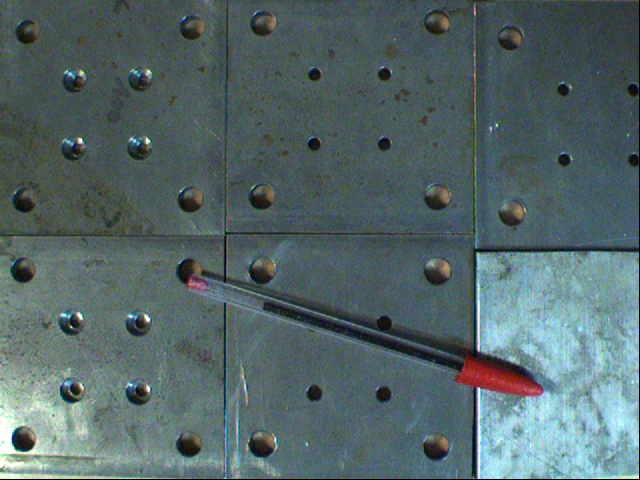}};
    \node[inner sep=0pt] (im10) at (9.55, -1.8)
        {\includegraphics[width=.15\textwidth]{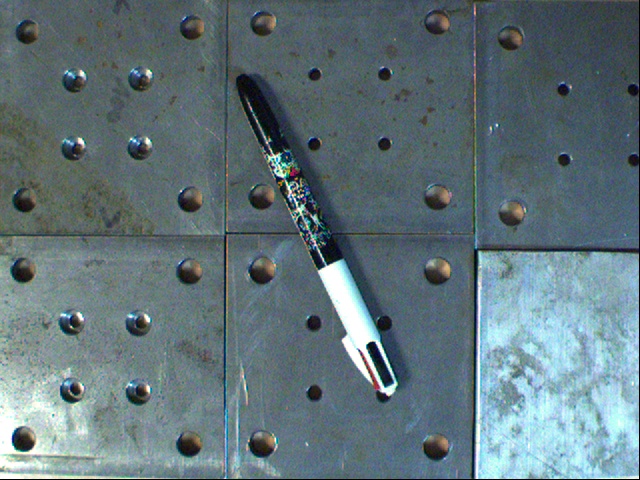}};
    \node[inner sep=0pt] (im11) at (11.9, -1.8)
        {\includegraphics[width=.15\textwidth]{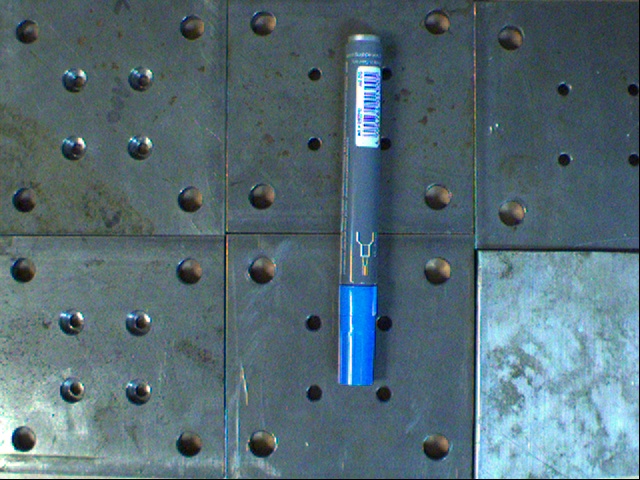}};
    \node[inner sep=0pt] (im12) at (0, -3.6)
        {\includegraphics[width=.15\textwidth]{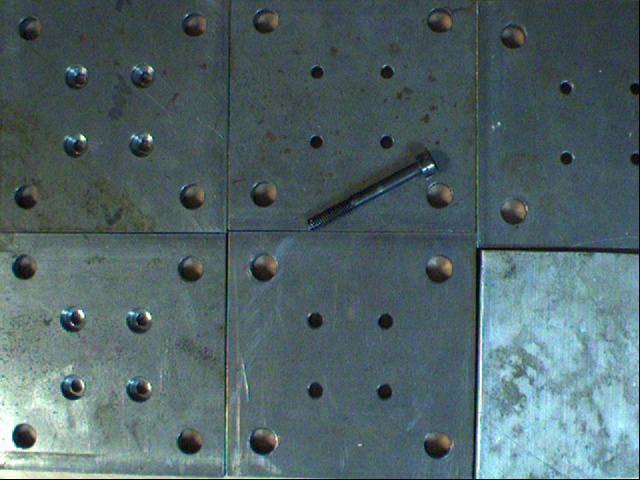}};
    \node[inner sep=0pt] (im13) at (2.35, -3.6)
        {\includegraphics[width=.15\textwidth]{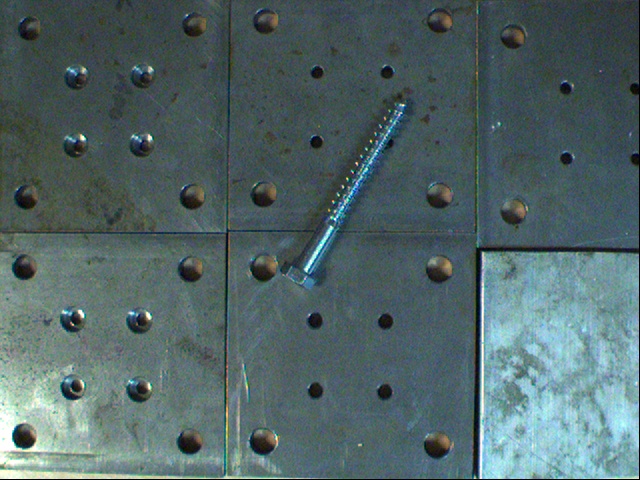}};
    \node[inner sep=0pt] (im14) at (4.7, -3.6)
        {\includegraphics[width=.15\textwidth]{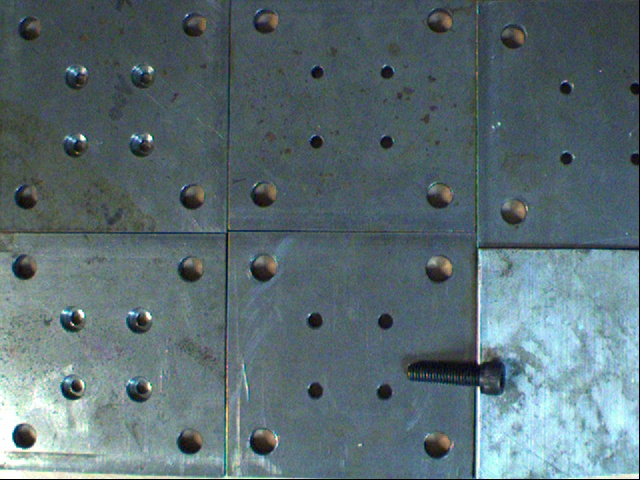}};
    \node[inner sep=0pt] (im15) at (7.2, -3.6)
        {\includegraphics[width=.15\textwidth]{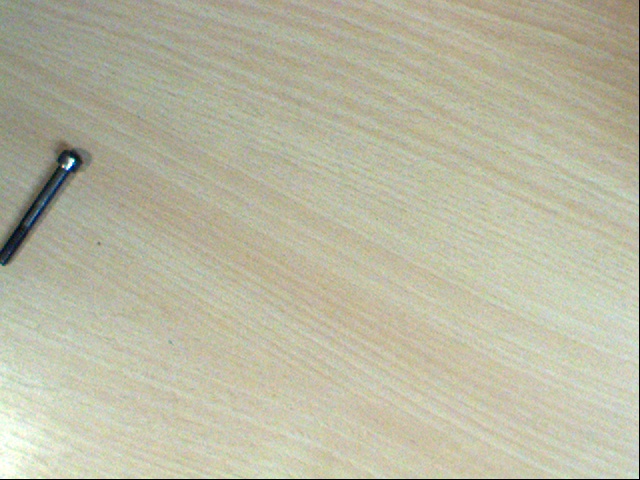}};
    \node[inner sep=0pt] (im16) at (9.55, -3.6)
        {\includegraphics[width=.15\textwidth]{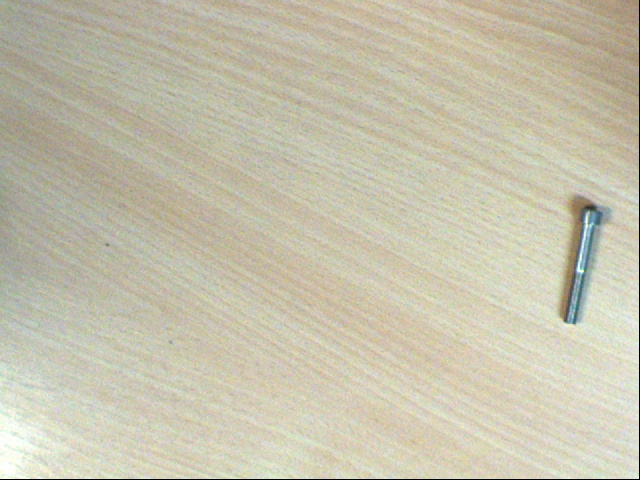}};
    \node[inner sep=0pt] (im17) at (11.9, -3.6)
        {\includegraphics[width=.15\textwidth]{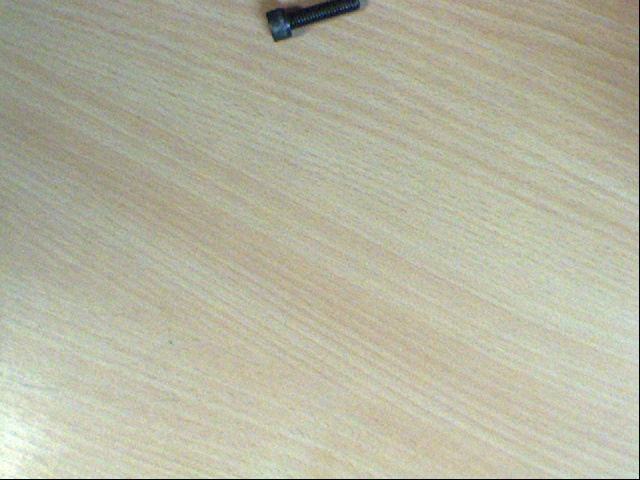}};
    \end{tikzpicture}
    
    \caption{Example images from the robustness validation dataset. Objects in the same row are expected to be clustered in the same group. The five different background/lighting conditions are represented in this figure.}
    \label{fig:dataset}
\end{figure}

For further testing of the application robustness, we have built a dataset for pixel-based object clustering. The full dataset, together with its description, can be found at (\url{https://github.com/jorisguerin/toolClustering_dataset}) and example images can be seen on Figure \ref{fig:dataset}. The dataset statistics are reported in table \ref{table_newdata}. This dataset is difficult because some classes have low intra-cluster similarity (usb) and extra-cluster similarity between some classes is relatively high (pens/screws). The lighting conditions as well as the background also change within the dataset, which makes the task even harder.

\input{table_new_dataset.tex}

For each conditions, we randomly pick one out of the several pictures (different position/orientation) of each object. The results reported in the top subtable of Table \ref{tab:appli} are averaged over $100$ random combinations. The clustering results are not perfect, looking at the misclassifications, the main source of error comes from classes with low intra-cluster similarity (pens, usb) and from background containing sharp edges (conditions 4 and 5).

We also carry out an experiment of fine grained classification with this dataset. Within each class, we try to group together the pictures representing the exact same object. Purity results can be found in the bottom subtable of Table \ref{tab:appli}. An interesting fact can be noticed about object recognition within a category. Classes responsible for decreasing the clustering quality in the first subtable are objects with the highest purity in the second table. Such remark makes sens as low intra-cluster similarity is good for the second task but harmful for the first one.

Looking at the results, this robustness validation dataset appears to be a challenging one for image-set clustering and could be used to validate further research in the field.

\input{table_appli.tex}

%% file: table_new_dataset.tex
\begin{table}[!ht]
\caption{Several key features about the constructed dataset.} 
\label{table_newdata}
\centering

    \begin{tabular}{c|c|c|c}
        Problem type & Image Size & \# Classes & \# Instances\tabularnewline \hline
        Object recognition & $640 \times 480$ & 7 & 560 \tabularnewline
    \end{tabular}

\vspace*{3pt}
   
\end{table}

%% file: table_appli.tex
\begin{table}[h]
\centering
\caption{Clustering accuracies for different background, lighting and orientation conditions on the tool clustering dataset.}
\label{tab:appli}

\begin{tabular}{c|c|c|c|c|c|c}
\multicolumn{7}{c}{Grouping by category}\vspace{1mm}\tabularnewline
\backslashbox{Metric}{Condition} & 1 & 2 & 3 & 4 & 5 & Mixed \\ \hline
Purity  & 0.85 & 0.94 & 0.84 & 0.69 & 0.79 & 0.58 \\ \hline
NMI score & 0.87 & 0.94 & 0.83 & 0.71 & 0.82 & 0.54
\end{tabular}

\vspace*{10pt}

\begin{tabular}{c|c|c|c|c|c}
\multicolumn{6}{c}{Recognizing object inside a category }\vspace{1mm}\tabularnewline
\backslashbox{Object}{Condition} & 1    & 2    & 3    & 4    & 5   \\ \hline
Allen  & 0.58 & 0.58 & 0.67 & 0.67 & 0.67 \\ \hline
Clamp  & 0.83 & 0.92 & 0.67 & 0.58 & 0.67 \\ \hline
Driver & 0.75 & 0.75 & 0.75 & 0.63 & 0.75 \\ \hline
Flat   & 0.58 & 0.83 & 0.67 & 0.58 & 0.58 \\ \hline
Pen    & 1.0  & 1.0  & 0.75 & 0.69 & 1.0  \\ \hline
Screws & 0.5  & 0.56 & 0.75 & 0.50 & 0.56 \\ \hline
USB    & 0.80 & 0.65 & 0.85 & 0.55 & 0.65
\end{tabular}
\end{table}

%% file: conclusion.tex
\section
  {Conclusion and perspectives}

\subsection
  {Conclusive remarks}

This paper extends the interesting work of \cite{feat_extrac1, feat_extrac2, feat_extrac3} about the transferability of CNN features. It shows that, even for unsupervised classification tasks, features extracted from deep CNN trained on large and diverse datasets, combined with classic clustering algorithms, can compete with more sophisticated and tuned image-set clustering methods. The fairly simple and naive pipeline proposed outperforms the best results reported in recent work, which raises the question of which research direction should be chosen to reach generic knowledge. Are efforts spent in developing image representation extractors more useful than simply building larger and more diverse datasets?

This approach is used to implement a robotic application using unsupervised image classification to store objects smartly. To validate this application, we also built a challenging dataset for image clustering that is made available to the research community.

\subsection
  {Future work}

The proposed improvements mainly go in the direction of the robotics application, which is still not robust enough to adapt perfectly to very different looking objects within a cluster and to difficult backgrounds and lighting conditions. If we want to make it work in difficult environments, the clustering pipeline needs to be improved. One possible direction is to tune the final clustering algorithm, indeed, the scikit clustering algorithms are used without any parameter tuning, setting hyperparameters to their default values.

The sorting application can also be improved by introducing automatic image segmentation, which would make it more suitable for practical uses. To do this, we could use a pretrained region proposal network \cite{roi_prop} and cluster objects in the proposed regions.